\documentclass[11pt,a4paper]{article}
\usepackage[hyperref]{acl2020}
\usepackage{times}
\usepackage{latexsym}

\usepackage{microtype}

\usepackage{tcolorbox}
\newtcbox{\orangebox}{on line,
  colframe=orange,colback=orange!10!white,
  boxrule=0.3mm,
  arc=1.0mm,
  boxsep=0pt,left=1pt,right=1pt,top=1pt,bottom=1pt
  }

\newtcbox{\greenbox}{on line,
  colframe=green!70!black,colback=green!10!white,
  boxrule=0.3mm,
  arc=1.0mm,
  boxsep=0pt,left=1.0pt,right=1.0pt,top=1.0pt,bottom=1.0pt
  }
  
\newtcbox{\blueboxa}{on line,
  colframe=blue!70!black,colback=blue!5!white,
  boxrule=0.3mm,
  arc=1.0mm,
  boxsep=0pt,left=1.2pt,right=1.2pt,top=1.3pt,bottom=1.3pt
  }

\newtcbox{\blueboxb}{on line,
  colframe=blue!70!black,colback=blue!5!white,
  boxrule=0.3mm,
  arc=1.0mm,
  boxsep=0pt,left=1.2pt,right=1.2pt,top=1.6pt,bottom=1.6pt
  }

\newtcbox{\graybox}{on line,
  colframe=white,colback=gray!15!white,
  boxrule=0mm,left=1pt,right=1pt,top=1pt,bottom=1pt
  }

\usepackage{xcolor}
\usepackage{tabularx}
\usepackage{amsmath}
\usepackage{amsfonts}
\usepackage{amssymb}

\usepackage{multirow}
\usepackage{here}
\usepackage{setspace}

\usepackage{tikz,tikz-qtree}
\usetikzlibrary{shapes.geometric,positioning,shapes,fit,arrows,backgrounds}
\usetikzlibrary{intersections, calc}
\usepackage{algorithm}
\usepackage{algorithmic}
\usepackage{proof}
\usepackage{ebproof}
\usepackage{textcomp}
\usepackage[T1]{fontenc}

\usepackage{url}
\usepackage{mediabb}
\usepackage{gb4ea}
\aclfinalcopy 

\newcommand{\reds}{red!80!black}
\newcommand{\blues}{blue!80!black}

\newcommand{\positive}[1]{\textit{#1}\,$\textcolor{\reds}{\uparrow}$}

\newcommand{\negative}[1]{\textit{#1}\,$\textcolor{\blues}{\downarrow}$}

\newcommand{\depth}[1]{\ensuremath{\textbf{D}_{#1}}}
\newcommand{\upq}{\ensuremath{q^{\uparrow}}}
\newcommand{\downq}{\ensuremath{q^{\downarrow}}}

\definecolor{salmon}{cmyk}{0,0.53,0.38,0}
\definecolor{cadetblue}{cmyk}{1,0.5,0,0}

\newcommand{\perm}{\ensuremath{\mathsf{perm}}}

\newcommand{\todo}[1]{\textcolor{black}{#1}} 

\newcommand{\adddata}{+}

\newcommand{\PredRepl}[1]{\underline{#1}}
\newcommand{\PredReplColor}[1]{\orangebox{#1}}
\newcommand{\Qf}[1]{\underline{#1}}
\newcommand{\QfColor}[1]{\greenbox{#1}}
\newcommand{\ClColora}[1]{\blueboxa{#1}}
\newcommand{\ClColorb}[1]{\blueboxb{#1}}

\newcommand{\Repl}[1]{\textsc{#1}}

\title{Do Neural Models Learn Systematicity \\
of Monotonicity Inference in Natural Language?}

\author{
  \parbox{\linewidth}{\centering
   Hitomi Yanaka$^1$,
   Koji Mineshima$^2$,
   Daisuke Bekki$^3$, and
   Kentaro Inui$^{4,1}$
  }
  \\
   $^1$\mbox{\rm RIKEN,}
   $^2$\mbox{\rm Keio University,}
   $^3$\mbox{\rm Ochanomizu University,}
   $^4$\mbox{\rm Tohoku University} 
  \\
  \parbox{\linewidth}{\centering
   {\tt hitomi.yanaka@riken.jp},
   {\tt minesima@abelard.flet.keio.ac.jp},
   {\tt bekki@is.ocha.ac.jp},
   {\tt inui@ecei.tohoku.ac.jp}
   }
}

\date{}
\begin{document}
\maketitle
\begin{abstract}
Despite the success of language models using neural networks, it remains unclear to what extent neural models have the generalization ability to perform inferences. In this paper, we introduce a method for evaluating whether neural models can learn \todo{\textit{systematicity} of monotonicity inference} in natural language, namely, the regularity for performing arbitrary inferences with generalization on composition. We consider four aspects of monotonicity inferences and test whether the models can systematically interpret \todo{lexical and logical phenomena} on different training/test splits. A series of experiments show that three neural models systematically draw inferences on unseen combinations of lexical and logical phenomena when the syntactic structures of the sentences are similar between the training and test sets. However, the performance of the models significantly decreases when the structures are slightly changed in the test set while retaining all vocabularies and constituents already appearing in the training set. This indicates that the generalization ability of neural models is limited to cases where the syntactic structures are nearly the same as those in the training set.

\end{abstract}
\section{Introduction}
\label{sec:intro}
Natural language inference (NLI), a task whereby a system judges whether given a set of premises $P$ semantically entails a hypothesis $H$~\cite{series/synthesis/2013Dagan,Bowman2015},
is a fundamental task for natural language understanding.
As with other NLP tasks, recent studies have shown a remarkable impact of deep neural networks in NLI~\cite{DBLP:journals/corr/WilliamsNB17,wang2018glue,BERT2018new}.
However, it remains unclear to what extent DNN-based models are capable of learning the compositional generalization underlying NLI from given labeled training instances. 

Systematicity of inference
(or \textit{inferential systematicity})
~\cite{Fodor1988-FODCAC,Aydede1997} in natural language has been intensively studied in the field of formal semantics.
From among the various aspects of inferential systematicity, in the context of NLI,
we focus on \emph{monotonicity}~\cite{10.2307/25001141,moss2014} and
its \emph{productivity}.
Consider the following premise--hypothesis pairs (1)--(3), which have the target label \textit{entailment}:

{\footnotesize
\begin{exe}
\ex \label{ex:1}
\begin{xlist}
\exi{$P$:} \label{ex:1a} \textit{\textbf{Some}} [\positive{\underline{puppies}}] \textit{ran}.
\exi{$H$:} \label{ex:1b} \textit{\textbf{Some} \underline{dogs} ran}.
\end{xlist}
\ex \label{ex:2}
\begin{xlist}
\exi{$P$:} \label{ex:2a} \textit{\textbf{No}} [\negative{\underline{cats}}] \textit{ran}.
\exi{$H$:}\label{ex:2b} \textit{\textbf{No} \underline{small cats} ran}.
\end{xlist}
\ex \label{ex:3}
\begin{xlist}
\exi{$P$:} \label{ex:3a} \textit{\textbf{Some}} [\textit{puppies which chased \textbf{no}} [\negative{\underline{cats}}]] \textit{ran}.
\exi{$H$:} \label{ex:3b} \textit{\textbf{Some} dogs which chased \textbf{no} \underline{small cats} ran}.
\end{xlist}
\end{exe}
}

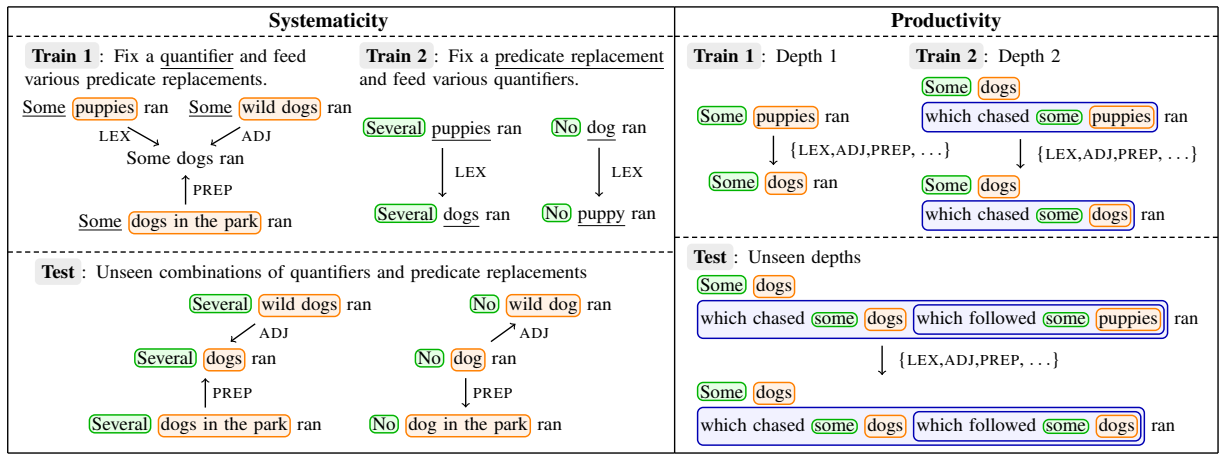
\begin{figure*}[h!tb]
\centering
\scalebox{0.65}{
\begin{tikzpicture}

\tikzset{headline1/.style={text width=6.5cm}};
\tikzset{headline2/.style={text width=12.5cm}};
\tikzset{sentence0/.style={text width=3.0cm, text centered}};
\tikzset{sentence1/.style={text width=4.0cm, text centered}};
\tikzset{sentence2/.style={text width=5.0cm, text centered}};
\tikzset{sentence3/.style={text width=7.5cm, text centered}};
\tikzset{sentence4/.style={text width=4.0cm}};
\tikzset{sentence5/.style={text width=7.5cm}};
\tikzset{dummy/.style={text width=0.5cm}};
\tikzset{entail/.style={color=black,thick,->}}


\draw[thick] (-6.0,0) -- (7.5,0);
\draw[thick] (-6.0,-9.1) -- (7.5,-9.1);
\draw[thick] (-6.0,-9.1) -- (-6.0,0);
\draw[thick] (7.5,-9.1) -- (7.5,0);

\draw[thick,densely dashed] (-6.0,-5) -- (7.5,-5);


\node (sys) at (0.5,-0.3) {\large\textbf{Systematicity}};

\draw[thick,densely dashed] (-6.0,-0.6) -- (7.5,-0.6);

\node[headline1] (train1) at (-2.4,-1.2) {\graybox{\textbf{Train 1}}: Fix a \PredRepl{quantifier} and feed \\various predicate replacements.};

\node[sentence1, below=1.0cm of train1] (train1H) {Some dogs ran};

\node[dummy, above=0.6cm of train1H] (train1above){};

\node[sentence0,left=-0.2cm of train1above] (train1P1)
{\Qf{Some} \PredReplColor{puppies} ran};
\node[sentence1,right=-0.8cm of train1above] (train1P2)
{\Qf{Some} \PredReplColor{wild dogs} ran};
\node[sentence2,below=0.6cm of train1H] (train1P3)
{\Qf{Some} \PredReplColor{dogs in the park} ran};

\draw[entail] (train1P1) edge node[left=3pt] {\Repl{lex}} (train1H);
\draw[entail] (train1P2) edge node[right=5pt] {\Repl{adj}} (train1H);
\draw[entail] (train1P3) edge node[right] {\Repl{prep}} (train1H);

\node[headline1,right=0cm of train1] (train2) {\graybox{\textbf{Train 2}}: Fix a \PredRepl{predicate replacement} \\and feed various quantifiers.};

\node[sentence1] at (2.8,-2.5) (train2P1) {\QfColor{Several} \PredRepl{puppies} ran};
\node[sentence0,below=1.0cm of train2P1] (train2H1) {\QfColor{Several} \PredRepl{dogs} ran};

\node[sentence0,right=-0.6cm of train2P1] (train2P2) {\QfColor{No} \PredRepl{dog} ran};
\node[sentence0,below=1.0cm of train2P2] (train2H2) {\QfColor{No} \PredRepl{puppy} ran};

\draw[entail] (train2P1) edge node[right=3pt] {\Repl{lex}} (train2H1);
\draw[entail] (train2P2) edge node[right=3pt] {\Repl{lex}} (train2H2);

\node[headline2] (test) at (0.8,-5.4) {\graybox{\textbf{Test}}: Unseen combinations of quantifiers and predicate replacements};

\node[sentence1] (test1H) at (-2.0,-7.2) {\QfColor{Several} \PredReplColor{dogs} ran};

\node[dummy, above=0.6cm of test1H] (test1above){};

\node[sentence1,right=-1.0cm of test1above] (test1P2)
{\QfColor{Several} \PredReplColor{wild dogs} ran};
\node[sentence2,below=0.6cm of test1H] (test1P3)
{\QfColor{Several} \PredReplColor{dogs in the park} ran};

\draw[entail] (test1P2) edge node[right=5pt] {\Repl{adj}} (test1H);
\draw[entail] (test1P3) edge node[right] {\Repl{prep}} (test1H);

\node[sentence1, right=1.0cm of test1H] (test2H) {\QfColor{No} \PredReplColor{dog} ran};

\node[dummy, above=0.6cm of test2H] (test2above){};

\node[sentence1,right=-1.0cm of test2above] (test2P2)
{\QfColor{No} \PredReplColor{wild dog} ran};
\node[sentence2,below=0.6cm of test2H] (test2P3)
{\QfColor{No} \PredReplColor{dog in the park} ran};

\draw[entail] (test2H) edge node[right=5pt] {\Repl{adj}} (test2P2);
\draw[entail] (test2H) edge node[right] {\Repl{prep}} (test2P3);

\draw[thick] (7.5,0) -- (18.5,0);
\draw[thick] (7.5,-9.1) -- (18.5,-9.1);
\draw[thick] (18.5,-9.1) -- (18.5,0);

\node (sys) at (13.0,-0.3) {\large\textbf{Productivity}};

\draw[thick,densely dashed] (7.5,-0.6) -- (18.5,-0.6);

\node[headline1] (train1d) at (11.0,-1.0) {\graybox{\textbf{Train 1}}: Depth 1};

\node[sentence1] (train1dH) at (9.5,-3.6) {\QfColor{Some} \PredReplColor{dogs} ran};

\node[sentence1,above=0.6cm of train1dH] (train1dP)
{\QfColor{Some} \PredReplColor{puppies} ran};

\draw[entail] (train1dP) edge node[right=5pt] {\{\Repl{lex},\Repl{adj},\Repl{prep}, $\ldots$\}} (train1dH);

\node[headline1] (train2d) at (15.5,-1.0) {\graybox{\textbf{Train 2}}: Depth 2};

\node[sentence4] (train2dH) at (14.5,-4.0) {\QfColor{Some} \PredReplColor{dogs} \ClColora{which chased \QfColor{some} \PredReplColor{dogs}}\mbox{ }ran};

\node[sentence4,above=0.6cm of train2dH] (train2dP)
{\QfColor{Some} \PredReplColor{dogs} \ClColora{which chased \QfColor{some} \PredReplColor{puppies}}\mbox{ }ran};

\draw[entail] (train2dP) edge node[right=5pt] {\{\Repl{lex},\Repl{adj},\Repl{prep}, $\ldots$\}} (train2dH);

\draw[thick,densely dashed] (7.5,-4.7) -- (18.5,-4.7);

\node[headline1] (testd) at (11.0,-5.1) {\graybox{\textbf{Test}}: Unseen depths};

\node[sentence5] (train2dH) at (11.7,-8.3) {\QfColor{Some} \PredReplColor{dogs} \ClColorb{which chased \QfColor{some} \PredReplColor{dogs} \ClColora{which followed \QfColor{some} \PredReplColor{dogs}}}\mbox{ }ran};

\node[sentence5,above=0.6cm of train2dH] (train2dP)
{\QfColor{Some} \PredReplColor{dogs} \ClColorb{which chased \QfColor{some} \PredReplColor{dogs} \ClColora{which followed \QfColor{some} \PredReplColor{puppies}}}\mbox{ }ran};

\draw[entail] (train2dP) edge node[right=5pt] {\{\Repl{lex},\Repl{adj},\Repl{prep}, $\ldots$\}} (train2dH);

\end{tikzpicture}
}

\caption{An illustration of the basic idea.
For Systematicity and Productivity, we train models on the \textbf{Train 1} and \textbf{Train 2} sets and test them on the \textbf{Test} set.
Arrow (\raisebox{0.7mm}{\tikz{\draw[thick,->](0,0) -- (0.4,0)}}) means entailment relation;
\Repl{lex}, \Repl{adj}, and \Repl{prep}
mean predicate replacements for
lexical relations, adjectives,
and prepositional phrases, respectively.
In Productivity, we use various quantifiers and predicate replacements in each depth.
}
\label{fig:goodpic}
\end{figure*}

\noindent As in (\ref{ex:1}), for example, quantifiers such as \textit{some} 
exhibit \textbf{upward monotone}
(shown as [...\positive{}]), and replacing a phrase in an upward-entailing context in a sentence with 
\todo{a more general phrase} (replacing \textit{puppies} in $P$ with \textit{dogs} as in $H$) yields a sentence inferable from the original sentence. 
In contrast, as in (\ref{ex:2}), quantifiers such as \textit{no} exhibit \textbf{downward monotone} (shown as [...\negative{}]), and replacing a phrase in a downward-entailing context with \todo{a more specific phrase} (replacing \textit{cats} in $P$ with \textit{small cats} as in $H$) yields a sentence inferable from the original sentence.
Such primitive inference patterns combine recursively as in (\ref{ex:3}).
This manner of \todo{monotonicity and its productivity} produces a potentially infinite number of inferential patterns.
Therefore, NLI models must be capable of systematically interpreting such primitive patterns and reasoning over unseen combinations of patterns.
\todo{Although many studies have addressed this issue by modeling logical reasoning in formal semantics~\cite{abzianidze:2015:EMNLP,D16-1242,Hu2019MonaLogAL} and testing DNN-based models on monotonicity inference~\cite{yanaka-etal-2019-neural,yanaka2019,Richardson2019},
the ability of DNN-based models to generalize to unseen combinations of patterns is still underexplored.}


\todo{Given this background, we investigate the systematic generalization ability of
DNN-based models on four aspects of monotonicity: 
(i)~systematicity of predicate replacements (i.e., replacements with a more general or specific phrase), (ii)~systematicity of embedding quantifiers, (iii)~productivity, and (iv)~localism (see Section~2.2).}
To this aim, we introduce a new evaluation protocol where we (i)~synthesize training instances from sampled sentences
and (ii)~systematically control which patterns are shown to the models in the training phase and which are left unseen.
The rationale behind this protocol is two-fold.
First, patterns of monotonicity inference are highly systematic, so we can create training data with arbitrary combinations of patterns, as in examples (\ref{ex:1})--(\ref{ex:3}). 
Second, evaluating the performance of the models trained with well-known NLI datasets such as MultiNLI~\cite{DBLP:journals/corr/WilliamsNB17} might severely underestimate the ability of the models because such datasets tend to contain only a limited number of training instances that exhibit the inferential patterns of interest. 
Furthermore, using such datasets would prevent us from identifying which combinations of patterns the models can infer from which patterns in the training data.

This paper makes two primary contributions.
First, we introduce an evaluation protocol\footnote{The evaluation code will be publicly available at https://github.com/verypluming/systematicity.} using the systematic control of the training/test split under various combinations of semantic properties to evaluate whether models learn inferential systematicity in natural language.
\todo{Second, we apply our evaluation protocol to three NLI models and present evidence suggesting that, while all models generalize to unseen combinations of lexical and logical phenomena, their generalization ability is limited to cases where sentence structures are nearly the same as those in the training set.}

\section{Method}
\subsection{Basic idea}
Figure~\ref{fig:goodpic} illustrates the basic idea of our evaluation protocol on monotonicity inference.
\todo{We use synthesized monotonicity inference datasets, where NLI models should capture both (i)~monotonicity directions (upward/downward) of various quantifiers and (ii)~the types of various predicate replacements in their arguments.
To build such datasets, we first generate a set of premises} $G_d^{\textbf{Q}}$ by a context-free grammar $G$ with depth $d$ (i.e., the maximum number of applications of recursive rules), given a set of quantifiers $\textbf{Q}$.
Then, by applying $G_d^{\textbf{Q}}$ to elements of a set of functions for predicate replacements (or \textit{replacement functions} for short) $\textbf{R}$ \todo{that rephrase a constituent in the input premise and return a hypothesis},
we obtain a set $\textbf{D}_d^{\textbf{Q},\textbf{R}}$ of premise--hypothesis pairs defined as

\fontsize{9.5pt}{0cm}\selectfont
\begin{align*}
  \textbf{D}_d^{\textbf{Q},\textbf{R}} =\;& \{(P, H) \mid \, P \in G_d^{\textbf{Q}},\;\exists r \in \textbf{R}\;\;(r(P) = H)\}.
\end{align*}
\normalsize

For example, the premise \textit{Some puppies ran} is generated from the quantifier \textit{some} in \textbf{Q} and the production rule $\textit{S}\rightarrow \textit{Q},\textit{N},\textit{IV}$, and thus it is an element of $G_1^{\textbf{Q}}$.
By applying this premise to a replacement function that replaces the word in the premise with its hypernym (e.g., $\textit{puppy} \sqsubseteq \textit{dog}$), we provide the premise--hypothesis pair $\textit{\textbf{Some} \underline{puppies} ran}\Rightarrow \textit{\textbf{Some} \underline{dogs} ran}$ in Fig.~\ref{fig:goodpic}.

We can control which patterns are shown to the models during training and which are left unseen by systematically splitting $\textbf{D}_d^{\textbf{Q},\textbf{R}}$ into training and test sets.
\todo{As shown on the left side of Figure~\ref{fig:goodpic}}, we consider how to test the systematic capacity of models with unseen combinations of quantifiers and predicate replacements.
To expose models to primitive patterns regarding \textbf{Q} and \textbf{R},
we fix an arbitrary element $q$ from $\textbf{Q}$
and feed various predicate replacements into the models from the training set of inferences $\textbf{D}_d^{\{q\}, \textbf{R}}$ generated from combinations of the fixed quantifier and all predicate replacements.
Also, we select an arbitrary element $r$ from $\textbf{R}$
and feed various quantifiers into the models from the training set
of inferences $\textbf{D}_d^{\textbf{Q}, \{r\}}$ generated from combinations of all quantifiers and the fixed predicate replacement.

We then test the models on the set of inferences generated from unseen combinations of quantifiers and predicate replacements.
That is, we test them on the set of inferences
$\textbf{D}_d^{\overline{\{q\}},\overline{\{r\}}}$ generated from the complements $\overline{\{q\}},\overline{\{r\}}$ of $\{q\}, \{r\}$.
\todo{If models capture inferential systematicity in combinations of quantifiers and predicate replacements, they can correctly perform all inferences in $\textbf{D}_d^{\overline{\{q\}},\overline{\{r\}}}$ on an arbitrary split based on $q,r$.}

\todo{Similarly, as shown on the right side of Figure~\ref{fig:goodpic}, we can test the productive capacity of models with unseen depths by changing the training/test split based on $d$.}
\todo{For example, by training models on $\textbf{D}_d^{\textbf{Q},\textbf{R}}$ and testing them on $\textbf{D}_{d+1}^{\textbf{Q},\textbf{R}}$,}
we can evaluate whether models generalize to one deeper depth.
By testing models
with an arbitrary training/test split of $\textbf{D}_d^{\textbf{Q},\textbf{R}}$ based on semantic properties of monotonicity inference \todo{(i.e., quantifiers, predicate replacements, and depths)},
we can evaluate whether models systematically interpret them.

\subsection{Evaluation protocol} 

\todo{To test NLI models from multiple perspectives of inferential systematicity in monotonicity inferences,
we
focus on four aspects: (i)~systematicity of predicate replacements, (ii)~systematicity of embedding quantifiers, (iii)~productivity, and (iv)~localism.
For each aspect, we use a set $\textbf{D}_d^{\textbf{Q},\textbf{R}}$ of premise--hypothesis pairs.
Let $\textbf{Q} = \textbf{Q}^{\uparrow} \cup \textbf{Q}^{\downarrow}$ be the union of a set of selected upward quantifiers $\textbf{Q}^{\uparrow}$ and a set of selected downward quantifiers $\textbf{Q}^{\downarrow}$ such that $|\textbf{Q}^{\uparrow}|\!=\!|\textbf{Q}^{\downarrow}|\!=\!n$.}
Let \textbf{R} be a set of replacement functions
$\{r_1, \ldots, r_m\}$, and
$d$ be the embedding depth, with $1 \leq d \leq s$.
(4)~is an example of an element of $\depth{1}^{\textbf{Q},\textbf{R}}$, 
containing the quantifier \textit{some} in the subject position and the predicate replacement \todo{using the hypernym relation} $\textit{dogs}\sqsubseteq\textit{animals}$ in its upward-entailing context without embedding.

\scalebox{0.9}{
\kern-2em
\begin{tabular}{llcl}
 (4)&$P$: \textit{\textbf{Some} \underline{dogs} ran}&$\Rightarrow$&
 $H$: \textit{\textbf{Some} \underline{animals} ran}\\
\end{tabular}
}

\paragraph{I. Systematicity of predicate replacements}
The following describes how we test the extent to which models generalize to unseen combinations of quantifiers and predicate replacements.
Here, we expose models to all primitive patterns of predicate replacements like (4) and (5) and
all primitive patterns of quantifiers like (6) and (7).
\todo{We then test whether the models can systematically capture the difference
between upward quantifiers (e.g., \textit{several}) and downward quantifiers (e.g., \textit{no}) as well as the different types of predicate replacements (e.g.,
the lexical relation
$\textit{dogs}\sqsubseteq\textit{animals}$
and the adjective deletion
$\textit{small dogs}\sqsubseteq\textit{dogs}$)} and correctly interpret unseen combinations of quantifiers and predicate replacements like (8) and (9).

\scalebox{0.75}{
\kern-2em
\begin{tabular}{llcl}
 (5)&$P$: \textit{\textbf{Some} \underline{small dogs} ran}&$\Rightarrow$&
 $H$: \textit{\textbf{Some} \underline{dogs} ran}\\
 (6)&$P$: \textit{\textbf{Several} \underline{dogs} ran}&$\Rightarrow$&
 $H$: \textit{\textbf{Several} \underline{animals} ran}\\
 (7)&$P$: \textit{\textbf{No} \underline{animals} ran}&$\Rightarrow$&
 $H$: \textit{\textbf{No} \underline{dogs} ran}\\
 (8)&$P$: \textit{\textbf{Several} \underline{small dogs} ran}&$\Rightarrow$&
 $H$: \textit{\textbf{Several} \underline{dogs} ran}\\
 (9)&$P$: \textit{\textbf{No} \underline{dogs} ran}&$\Rightarrow$&
 $H$: \textit{\textbf{No} \underline{small dogs} ran}\\
\end{tabular}
}

Here, we consider a set of inferences $\depth{1}^{\textbf{Q},\textbf{R}}$ whose depth is 1. 
We move from harder to easier tasks by gradually changing the training/test split according to combinations of quantifiers and predicate replacements.
First, we expose models to primitive patterns of \textbf{Q} and \textbf{R} with the minimum training set.
Thus, we define the initial training set $\textbf{S}_1$ and test set $\textbf{T}_1$ as follows:
\begin{align*}
  (\textbf{S}_1, \textbf{T}_1) =\;& (\textbf{D}_1^{\{q\},\textbf{R}} \cup \textbf{D}_1^{\textbf{Q},\{r\}},\ \textbf{D}_1^{\overline{\{q\}},\overline{\{r\}}})
\end{align*}
where $q$ is arbitrarily selected from \textbf{Q}, and $r$ is arbitrarily selected from \textbf{R}.

Next, we gradually add the set of inferences generated from combinations of an upward--downward quantifier pair and all predicate
replacements to the training set.
In the examples above, we add (8) and (9) to the training set to simplify the task.
We assume a set $\textbf{Q}'$ of a pair of upward/downward quantifiers, namely, $\{(\upq,\downq) \mid (\upq,\downq)\subseteq \textbf{Q}^{\uparrow} \times \textbf{Q}^{\downarrow},\ \upq,\downq \neq q\}$.
We consider a set $\perm(\textbf{Q}')$ consisting of permutations of $\textbf{Q}'$.
For each $p \in \perm(\textbf{Q}')$, we gradually add a set of inferences generated from $p(i)$ to the training set $\textbf{S}_i$ 
with
$1 < i \leq n-1$.
Then, we 
provide a test set $\textbf{T}_i$ generated from the complement $\overline{\textbf{Q}_i}$ of $\textbf{Q}_i =   \{x\mid\exists y (x,y) \in \textbf{Q}_i' \; \text{or} \; \exists y (y,x) \in \textbf{Q}_i' \}$ and $\overline{\{r\}}$ where $\textbf{Q}_i' = \{p(1), \ldots, p(i)\}$.
This protocol is summarized as
\begin{align*}
  \textbf{S}_{i+1} =\;& \textbf{S}_i \cup \depth{1}^{ \{\upq_i,\downq_i\},\textbf{R}}, \\
  \textbf{T}_i =\;&\depth{1}^{\overline{\textbf{Q}_i},\overline{\{r\}}} \quad \text{with} \; 1 <i \leq n-1
\end{align*}
where $(\upq_i,\downq_i)=p(i)$.

To evaluate the extent to which the generalization ability of models is robust for different syntactic structures, we use an additional test set $\textbf{T}'_i = \depth{1}^{\overline{\textbf{Q}_i},\overline{\{r\}}}$ generated using three production rules.
The first is the case where one adverb is added at the beginning of the sentence,
as in example~(\ref{ex:adv}).

\setcounter{exx}{9}
{\footnotesize
\begin{exe}
\ex \label{ex:adv}
\begin{xlist}
    \exi{$P_{adv}$:} \textit{Slowly, \textbf{several} \underline{small dogs} ran}
    \exi{$H_{adv}$:} \textit{Slowly, \textbf{several} \underline{dogs} ran}
\end{xlist}
\end{exe}
}

\noindent The second is the case where a three-word prepositional phrase is added at the beginning of the sentence, as in example~(\ref{ex:prep}).

{\footnotesize
\begin{exe}
\ex \label{ex:prep}
\begin{xlist}
    \exi{$P_{prep}$:} \textit{Near the shore, \textbf{several} \underline{small dogs} ran}
    \exi{$H_{prep}$:} \textit{Near the shore, \textbf{several} \underline{dogs} ran}
 \end{xlist}
\end{exe}
}

\noindent The third is the case where the replacement is performed in the object position, as in example~(\ref{ex:obj}).

{\footnotesize
\begin{exe}
\ex \label{ex:obj}
 \begin{xlist}
    \exi{$P_{obj}$:} \textit{Some tiger touched \textbf{several} \underline{small dogs}}
    \exi{$H_{obj}$:} \textit{Some tiger touched \textbf{several} \underline{dogs}}
 \end{xlist}
\end{exe}
}

\noindent We train and test models $|\perm(\textbf{Q}')|$ times, then take the average accuracy as the final evaluation result.
\begin{table*}[tb]
\centering
\scalebox{0.90}{
{\footnotesize
\begin{tabular}{ l l l l p{30em} c } \hline
\multicolumn{1}{c}{\textbf{Depth}} & \multicolumn{1}{c}{\textbf{Pred.}} &\multicolumn{1}{c}{\textbf{Monotone}} &\multicolumn{1}{c}{\textbf{Arg.}} &\multicolumn{1}{c}{ \textbf{Example }(premise, hypothesis, label)} & 
\multicolumn{1}{c}{\textbf{Avg. Len.}} \\ \hline
1&\texttt{CONJ}&\texttt{DOWNWARD}&\texttt{SECOND}& \begin{tabular}{p{30em}} \emph{\textbf{Less than three} lions \underline{left}.} \\
\emph{\textbf{Less than three} lions \underline{left and cried}.}\ \texttt{ENTAILMENT}
\end{tabular} & 
4.6 \\ \hline
2&\texttt{PP}&\texttt{UPWARD}&\texttt{FIRST}& \begin{tabular}{p{30em}} \emph{\textbf{Few} lions that hurt \textbf{at most three} \underline{small dogs} walked.} \\
\emph{\textbf{Few} lions that hurt \textbf{at most three} \underline{dogs} walked.}\ \texttt{ENTAILMENT}
\end{tabular} & 
9.0 \\ \hline
3&\texttt{AdJ}&\texttt{DOWNWARD}&\texttt{FIRST}& \begin{tabular}{p{30em}} \emph{\textbf{Some} elephant \textbf{no} rabbit which touched \textbf{a few} \underline{dogs} hit rushed.} \\
\emph{\textbf{Some} elephant \textbf{no} rabbit which touched \textbf{a few} \underline{small dogs} hit rushed.}\ \texttt{ENTAILMENT}
\end{tabular} & 
12.3 \\ \hline
4&\texttt{RC}&\texttt{UPWARD}&\texttt{FIRST}& \begin{tabular}{p{30em}} \emph{\textbf{Less than three} tigers which accepted \textbf{several} rabbits that loved \textbf{several} foxes \textbf{more than three} \underline{monkeys} cleaned dawdled.} \\
\emph{\textbf{Less than three} tigers which accepted \textbf{several} rabbits that loved \textbf{several} foxes \textbf{more than three} \underline{monkeys which ate dinner} cleaned dawdled.}\ \texttt{ENTAILMENT}
\end{tabular} &
16.6 \\ \hline
\end{tabular}}}

\caption{Examples of generated premise--hypothesis pairs. Depth: depth of embedding; Pred.: type of predicate replacements; Monotone: direction of monotonicity; Arg.: argument where the predicate replacement is performed; Avg. Len.: average sentence length.}
\label{tab:examples}
\end{table*}

\paragraph{I\hspace{-.1em}I. Systematicity of embedding quantifiers}
To properly interpret embedding monotonicity, models should detect both (i)~the monotonicity direction of each quantifier and (ii)~the type of predicate replacements in the embedded argument.
The following describes how we test whether models generalize to unseen combinations of embedding quantifiers.
We expose models to all primitive combination patterns of quantifiers and predicate replacements like (4)--(9) with a set of non-embedding monotonicity inferences $\depth{1}^{\textbf{Q},\textbf{R}}$
and some embedding patterns like (\ref{ex:12}), where \textbf{\textit{Q}$_1$} and \textbf{\textit{Q}$_2$} are chosen from a selected set of upward or downward quantifiers such as \textit{some} or \textit{no}.
We then test the models with an inference with an unseen quantifier \textit{several} in (\ref{ex:16}) to evaluate whether models can systematically interpret embedding quantifiers.
\begin{exe} \small
\ex \label{ex:12}
\begin{xlist}
\exi{$P$:} \textit{\textbf{Q$_1$} animals that chased  \textbf{Q$_2$} \underline{dogs} ran}
\exi{$H$:} \textit{\textbf{Q$_1$} animals that chased 
\textbf{Q$_2$} \underline{animals} ran}
\end{xlist}
\ex \label{ex:16}
\begin{xlist}
\exi{$P$:}  \hspace{-1ex}\textit{\textbf{Several} animals that chased \textbf{several} \underline{dogs} ran}
\exi{$H$:} \hspace{-1ex}\textit{\textbf{Several} animals that chased \textbf{several} \underline{animals} ran}
\end{xlist}
\end{exe}
We move from harder to easier tasks of learning embedding quantifiers by gradually changing the training/test split of a set of inferences $\depth{2}^{\textbf{Q},\textbf{R}}$ whose depth is 2, i.e., inferences involving one embedded clause.

We assume a set $\textbf{Q}'$ of a pair of upward and downward quantifiers as $\textbf{Q}' \equiv \{(\upq,\downq)\mid(\upq,\downq)\subseteq \mathbf{Q}^{\uparrow}\times \mathbf{Q}^{\downarrow}\}$, and consider a set $\perm(\textbf{Q}')$ consisting of permutations of $\textbf{Q}'$.
For each $p \in \perm(\textbf{Q}')$, we gradually add a set of inferences $\depth{2}$ generated from $p(i)$ to the training set $\textbf{S}_i$ with $1\leq i \leq n-1$.

We test models trained with $\textbf{S}_i$ on a test set $\textbf{T}_i$ generated from the complement $\overline{\textbf{Q}_i}$ of $\textbf{Q}_i = \{x\mid\exists y (x,y) \in \textbf{Q}_i' \; \text{or} \; \exists y (y,x) \in \textbf{Q}_i' \}$ where $\textbf{Q}_i' = \{p(1), \ldots, p(i)\}$, summarized as
\begin{align*}
  \textbf{S}_0=\;& \textbf{D}_1^{\textbf{Q},\textbf{R}},\\
  \textbf{S}_i=\;& \textbf{S}_{i-1} \cup \depth{2}^{ \{\upq_i,\downq_i\},\textbf{R}},\\
  \textbf{T}_i=\;&\depth{2}^{ \overline{\textbf{Q}_i},\textbf{R}} \quad \text{with} \; 1 \leq i \leq n-1
\end{align*}
where $(\upq_i,\downq_i)=p(i)$.
We train and test models $|\perm(\textbf{Q}')|$ times, then take the average accuracy as the final evaluation result.

\paragraph{I\hspace{-.1em}I\hspace{-.1em}I. Productivity} Productivity \todo{(or \textit{recursiveness})} is a concept related to systematicity, which refers to the capacity to grasp an indefinite number of natural language sentences or thoughts \todo{with generalization on composition}.
The following describes how we test whether models generalize to unseen deeper depths in embedding monotonicity
(see also the right side of Figure~\ref{fig:goodpic}).
For example, we expose models to all primitive non-embedding/single-embedding patterns like (\ref{ex:17}) and (\ref{ex:18}) and then test them with deeper embedding patterns like (\ref{ex:19}).

{\footnotesize
\begin{exe}
\ex \label{ex:17}
\begin{xlist}
\exi{$P$:} \textit{\textbf{Some} \underline{dogs} ran}
\exi{$H$:} \textit{\textbf{Some} \underline{animals} ran}
\end{xlist}
\ex \label{ex:18}
\begin{xlist}
\exi{$P$:} \textit{\textbf{Some} animals which chased \textbf{some} \underline{dogs} ran}
\exi{$H$:} \textit{\textbf{Some} animals which chased \textbf{some} \underline{animals} ran}
\end{xlist}
\ex \label{ex:19}
\begin{xlist}
\exi{$P$:} \textit{\textbf{Some} animals which chased \textbf{some} cats which followed \textbf{some} \underline{dogs} ran}
\exi{$H$:} \textit{\textbf{Some} animals which chased \textbf{some} cats which followed \textbf{some} \underline{animals} ran}
\end{xlist}
\end{exe}
}

\noindent To evaluate models on the set of inferences involving embedded clauses with depths exceeding those in the training set,
we train models with $\bigcup_{d \in \{1,\ldots, i+1\}}\depth{d}$, where we refer to $\depth{d}^{\textbf{Q},\textbf{R}}$ as $\depth{d}$ for short,
and test the models on 
$\bigcup_{d \in \{i+2,\ldots, s\}}\depth{d}$ with $1 \leq i \leq s-2$.

\paragraph{I\hspace{-.1em}V. Localism}
According to the principle of compositionality, the meaning of a complex expression derives from the meanings of its constituents and how they are combined.
One important concern is how local the composition operations should be~\cite{Pagin2010-PAGCID}.
We therefore test whether models trained with inferences involving embedded monotonicity locally perform inferences composed of smaller constituents.
Specifically, we train models with examples like (\ref{ex:19}) and then test the models with examples like (\ref{ex:17}) and (\ref{ex:18}).
We train models with  
\depth{d} and test the models on 
$\bigcup_{k \in \{1,\ldots, d\}}\depth{k}$
with $3 \leq d \leq s$ .

\section{Experimental Setting}
\subsection{Data creation}
\label{sec:data}
To prepare the datasets shown in Table~\ref{tab:examples}, we first generate premise sentences involving quantifiers from a set of context-free grammar (CFG) rules and lexical entries, shown in Table~\ref{tab:lexicon}
in the Appendix.
We select 10 words from among nouns, intransitive verbs, and transitive verbs as lexical entries.
\todo{A set of quantifiers \textbf{Q} consists of eight elements;} we use a set of four downward quantifiers $\textbf{Q}^{\downarrow}=$\{\textit{no, at most three, less than three, few}\} and
a set of four upward quantifiers $\textbf{Q}^{\uparrow}=$\{\textit{some, at least three, more than three, a few}\}, which have the same monotonicity directions in the first and second arguments.
We thus consider $n\!=\!|\textbf{Q}^{\uparrow}|\!=\!|\textbf{Q}^{\downarrow}|\!=\!4$ \todo{in the protocol in Section 2.2}.
The ratio of each monotonicity direction (upward/downward) of generated sentences is set to $1:1$.
We then generate hypothesis sentences by applying replacement functions to premise sentences according to the polarities of constituents.
The set of replacement functions $\textbf{R}$ is composed of
the seven types of lexical replacements and phrasal additions 
in Table~\ref{tab:replace}.
We remove unnatural premise--hypothesis pairs in which the same words or phrases appear more than once.

\begin{table}[tb]
    \centering
    \scalebox{0.91}{
    \begin{tabular}{ll}\\ \hline
    \textbf{Function}&\textbf{Example}\\ \hline
         $r_1$: hyponym& \textit{dogs} $\sqsubseteq$ \textit{animals} \\
         $r_2$: adjective&\textit{small dogs} $\sqsubseteq$ \textit{dogs} \\
         $r_3$: preposition&\textit{dogs in the park} $\sqsubseteq$ \textit{dogs}\\
         $r_4$: relative clause&\textit{dogs which ate dinner} $\sqsubseteq$ \textit{dogs}\\
         $r_5$: adverb&\textit{ran quickly} $\sqsubseteq$ \textit{ran}\\
         $r_6$: disjunction&\textit{ran} $\sqsubseteq$ \textit{ran or walked}\\
         $r_7$: conjunction&\textit{ran and barked} $\sqsubseteq$ \textit{ran}\\ \hline
    \end{tabular}
    }
    \caption{Examples of replacement functions.}
    \label{tab:replace}
\end{table}

For embedding monotonicity, we 
consider inferences involving four types of replacement functions in the first argument of the quantifier in Table~\ref{tab:replace}: hyponyms, adjectives, prepositions, and relative clauses. 
We generate sentences up to the depth $d=5$.
\todo{There are various types of embedding monotonicity, including relative clauses, conditionals, and negated clauses.}
In this paper, we consider three types of embedded clauses: peripheral-embedding clauses and two kinds of center-embedding clauses, shown in Table~\ref{tab:lexicon}
in the Appendix.

The number of generated sentences exponentially increases with the depth of embedded clauses.
Thus, 
we limit the number of inference examples to 320,000, split into 300,000 examples for the training set and 20,000 examples for the test set. 
We guarantee that all combinations of quantifiers are included in the set of inference examples for each depth.
Gold labels for generated premise--hypothesis pairs are automatically determined according to the polarity of the
argument position (upward/downward) and the type of predicate replacements (with more general/specific phrases).
The ratio of each gold label (entailment/non-entailment) in the training and test sets is set to $1:1$.

\begin{figure*}[tb]
\begin{center}
\scalebox{0.41}{\includegraphics[bb=0.000000 0.000000 1152.002304 288.000576]{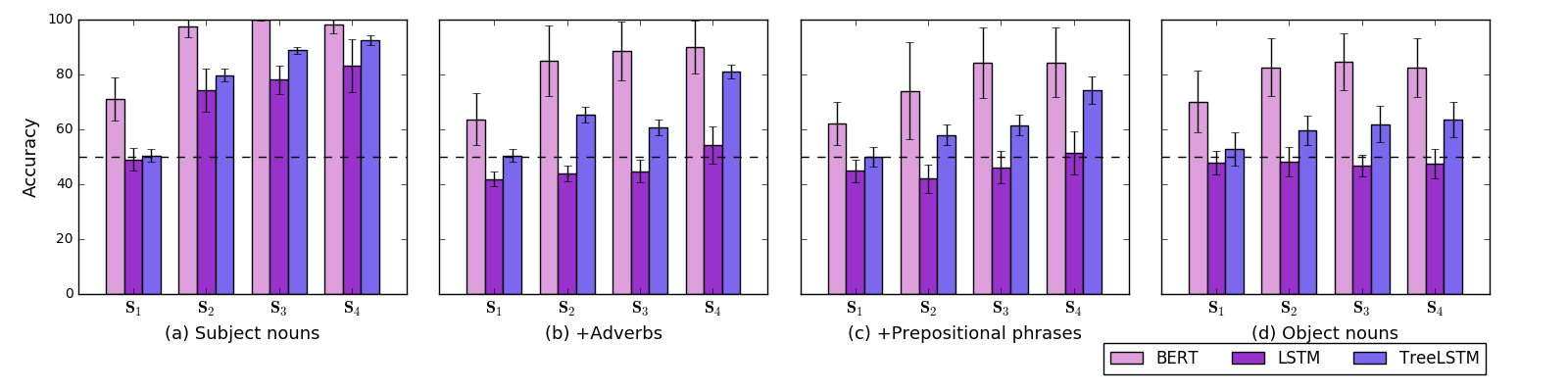}}
\end{center}
 \caption{Results for systematicity of predicate replacements.
 Accuracy on test sets where (a)~the replacement is performed in the subject position, (b)~one adverb is added at the beginning of the sentence, (c)~one three-word prepositional phrase is added at the beginning of the sentence, and (d)~the replacement is in the object position. $\textbf{S}_n$ indicates the experimental setting where the training set $\textbf{S}_n$ is used.}
 \label{fig:lex1_4}
\end{figure*}

To double-check the gold label, we translate each premise--hypothesis pair into a logical formula (see the Appendix for more details).
\todo{The logical formulas are
obtained by combining lambda terms in accordance with meaning composition rules specified in the CFG rules in the standard way~\cite{BlackburnBos05}.}
We prove the entailment relation using the theorem prover Vampire\footnote{https://github.com/vprover/vampire},
checking whether a proof is found in time for each entailment pair.
\todo{For all pairs, the output of the prover matched with the entailment relation automatically determined by monotonicity calculus.}

\subsection{Models}
We consider three
DNN-based NLI models.
The first architecture
employs long short-term memory (LSTM) networks~\cite{Hochreiter:1997:LSM:1246443.1246450}.
We set the number of layers to three with no attention.
Each premise and hypothesis is processed as a sequence of words using a recurrent neural network with LSTM cells, and the final hidden state of each serves as its representation.

The second architecture employs multiplicative tree-structured LSTM (TreeLSTM) networks~\cite{tran-cheng-2018-multiplicative}\todo{, which are expected to be more sensitive to 
hierarchical syntactic structures}.
Each premise and hypothesis is processed as a tree structure by bottom-up combinations of constituent nodes using the same shared compositional function, input word information, and between-word relational information.
\todo{We parse all premise--hypothesis pairs with the dependency parser using the spaCy library\footnote{https://spacy.io/} and obtain tree structures.
For each experimental setting, we randomly sample 100 tree structures and check their correctness.}
In LSTM and TreeLSTM, the dimension of hidden units is 200, and we initialize the word embeddings with
300-dimensional GloVe vectors~\cite{pennington-etal-2014-glove}.
Both models are optimized with Adam~\cite{Adam}, and no dropout is applied.

The third architecture is a Bidirectional Encoder Representations from Transformers (BERT) model~\cite{BERT2018new}.
We used the base-uncased model pre-trained on Wikipedia and BookCorpus from the pytorch-pretrained-bert library\footnote{https://github.com/huggingface/pytorch-pretrained-bert}, fine-tuned for the NLI task using our dataset. 
In fine-tuning BERT, no dropout is applied,
\todo{and we choose hyperparameters that are commonly used for MultiNLI.}
We train all models over 25 epochs or until convergence, and select the best-performing model based on its performance on the validation set.
We perform five runs per model and report the average and standard deviation of their scores.

\section{Experiments and Discussion}
\paragraph{I. Systematicity of predicate replacements}
\label{sec:lex}
Figure~\ref{fig:lex1_4} shows the performance on unseen combinations of quantifiers and predicate replacements.
In the minimal training set $\textbf{S}_1$, the accuracy of LSTM and TreeLSTM was almost the same as chance, but that of BERT was around 75\%, suggesting that only BERT generalized to unseen combinations of quantifiers and predicate replacements.
When we train BERT with the training set $\textbf{S}_2$, which contains inference examples generated from combinations of one pair of upward/downward quantifiers and all predicate replacements, the accuracy was 100\%.
This indicates that
\todo{by being taught two kinds of quantifiers in the training data, BERT could distinguish between upward and downward for the other quantifiers.}
The accuracy of LSTM and TreeLSTM increased with increasing the training set size, but did not reach 100\%.
This indicates that LSTM and TreeLSTM
also generalize to inferences involving similar quantifiers to some extent, but their generalization ability is imperfect.

When testing models with inferences where adverbs or prepositional phrases are added to the beginning of the sentence, the accuracy of all models 
significantly decreased.
This decrease becomes larger as the syntactic structures of the sentences in the test set become increasingly different from those in the training set.
\todo{
Contrary to our expectations, the models fail to maintain accuracy on test sets whose difference from the training set is the structure with the adverb at the beginning of a sentence.
Of course, we could augment datasets involving that structure, but doing so would require feeding all combinations of inference pairs into the models.}
These results indicate that the models tend to estimate the entailment label from the beginning of a premise--hypothesis sentence pair, and that inferential systematicity to draw inferences involving quantifiers and predicate replacements is not completely generalized at the level of arbitrary constituents.

\begin{figure}[tb]
 \scalebox{0.40}{\includegraphics[bb=0.000000 0.000000 576.001152 288.000576]{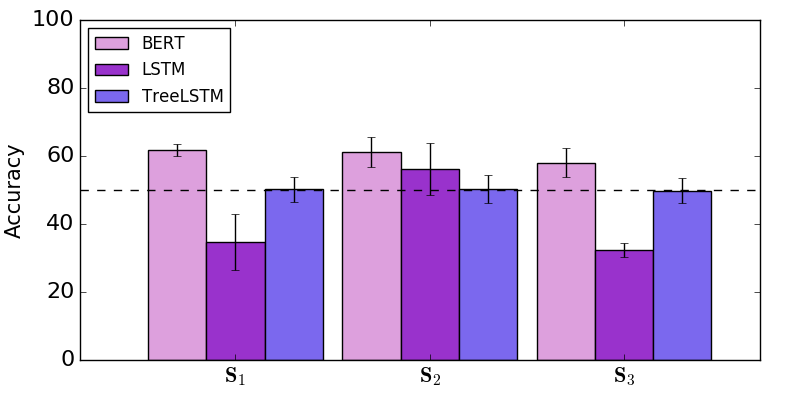}}
 \caption{Results for systematicity of embedding quantifiers.
 $\textbf{S}_n$ indicates the experimental setting where the training set $\textbf{S}_n$ is used.}
 \label{fig:emb3_6}
\end{figure}

\paragraph{I\hspace{-.1em}I. Systematicity of embedding quantifiers}
\label{sec:embq}
Figure~\ref{fig:emb3_6} shows the performance of all models on unseen combinations of embedding quantifiers.
Even when adding the training set of inferences involving one embedded clause and two quantifiers step-by-step, no model showed improved performance.
The accuracy of BERT slightly exceeded chance, but the accuracy of LSTM and TreeLSTM was nearly the same as or lower than chance.
These results suggest that 
all the models fail to generalize to unseen combinations of embedding quantifiers even when they involve similar upward/downward quantifiers.

\begin{table}[tb]
\centering
\scalebox{0.7}{
\begin{tabular}[t]{llccc} \hline
\textbf{Train} & \textbf{Dev/Test} & \textbf{BERT} & \textbf{LSTM} & \textbf{TreeLSTM} \\ \hline
$\depth{1}\adddata\depth{2}$& \depth{1} & 100.0$\pm$0.0 & 100.0$\pm$0.0 & 100.0$\pm$0.1  \\
 & \depth{2} & 100.0$\pm$0.0 & 99.8$\pm$0.2 & 99.5$\pm$0.1  \\
 & \depth{3} & 75.2$\pm$10.0 & 75.4$\pm$10.8 & 86.4$\pm$4.1  \\
 & \depth{4} & 55.0$\pm$3.7 & 57.7$\pm$8.7 & 58.6$\pm$7.8  \\
 & \depth{5} & 49.9$\pm$4.4 & 45.8$\pm$4.0 & 48.4$\pm$3.7  \\
 & \depth{3} (down) & $71.2\pm$4.0 & 70.4$\pm$4.0 & 86.4$\pm$4.1 \\
 & \depth{3} (up) & 80.5$\pm$7.5 & 84.7$\pm$4.9 & 86.4$\pm$4.1
 \\\hline
$\depth{1}\adddata\depth{2}\adddata\depth{3}$ & \depth{1} & 100.0$\pm$0.0 & 100.0$\pm$0.0 & 100.0$\pm$0.0  \\
 & \depth{2} & 100.0$\pm$0.0 & 95.1$\pm$7.8 & 99.6$\pm$0.0  \\
 & \depth{3} & 100.0$\pm$0.0 & 85.2$\pm$8.9 & 97.7$\pm$1.1  \\
 & \depth{4} & 77.9$\pm$10.8 & 59.7$\pm$10.8 & 68.0$\pm$5.6  \\
 & \depth{5} & 53.5$\pm$19.6 & 55.1$\pm$8.2 & 49.6$\pm$4.3  \\
 & \depth{4} (down) & $85.8\pm$10.5 & 76.9$\pm$6.6 & 68.0$\pm$5.6 \\
 & \depth{4} (up) & 86.8$\pm$1.8 & 81.1$\pm$5.6 & 68.0$\pm$5.6 \\
 \hline
 \end{tabular}
 }
  \caption{Results for productivity.
  $\depth{d}$ indicates the set of inferences where the embedding depth is $d$.}
 \label{tab:emb1_2}
\end{table}

\paragraph{I\hspace{-.1em}I\hspace{-.1em}I. Productivity}
\label{sec:depth}
Table~\ref{tab:emb1_2} shows the performance on unseen depths of embedded clauses.
\todo{The accuracy on \depth{1} and \depth{2} was nearly 100\%, indicating that all models almost completely generalize to inferences containing previously seen depths.}
When $\depth{1} \adddata \depth{2}$ were used as the training set, the accuracy of all models on \depth{3} exceeded chance.
Similarly, when $\depth{1} \adddata \depth{2} \adddata \depth{3}$ were used as the training set, the accuracy of all models on \depth{4} exceeded chance.
\todo{This indicates that all models partially generalize to inferences containing embedded clauses one level deeper than the training set.}

However, standard deviations of BERT and LSTM were around 10, suggesting that these models did not consistently generalize to inferences containing embedded clauses one level deeper than the training set.
While the distribution of monotonicity directions (upward/downward) in the training and test sets was uniform,
the accuracy of LSTM and BERT tended to be smaller for downward inferences than for upward inferences.
This also indicates that these models fail to properly compute monotonicity directions of constituents from syntactic structures.
The standard deviation of TreeLSTM was smaller, indicating that \todo{TreeLSTM robustly learns}
inference patterns containing embedded clauses one level deeper than the training set.

However, the performance of all models trained with $\depth{1} \adddata \depth{2}$ on \depth{4} and \depth{5} significantly decreased.
Also, performance decreased for all models trained with $\depth{1} \adddata \depth{2} \adddata \depth{3}$ on \depth{5}.
Specifically, there was significantly decreased performance of all models, including TreeLSTM, on inferences containing embedded clauses two or more levels deeper than those in the training set.
These results indicate that all models fail to develop productivity on inferences involving embedding monotonicity.

\paragraph{I\hspace{-.1em}V. Localism}
\label{sec:decomp}
\begin{table}[tb]
\centering
\scalebox{0.80}{
\begin{tabular}[t]{llccc} \hline
\textbf{Train} & \textbf{Dev/Test} & \textbf{BERT} & \textbf{LSTM} & \textbf{TreeLSTM} \\ \hline
\depth{3} & \depth{1} & 49.6$\pm$0.5 & 48.8$\pm$13.2 & 49.8$\pm$4.1  \\
 & \depth{2} & 49.8$\pm$0.6 & 47.3$\pm$12.1 & 51.8$\pm$1.1  \\
 & \depth{3} & 100.0$\pm$0.0 & 100.0$\pm$0.0 & 100.0$\pm$0.2  \\ \hline
\depth{4} & \depth{1} & 50.3$\pm$1.0 & 46.8$\pm$6.5 & 49.0$\pm$0.4  \\
 & \depth{2} & 49.6$\pm$0.8 & 45.4$\pm$1.8 & 49.7$\pm$0.3  \\
 & \depth{3} & 50.2$\pm$0.7 & 45.1$\pm$0.6 & 50.5$\pm$0.7  \\
 & \depth{4} & 100.0$\pm$0.0 & 100.0$\pm$0.0 & 100.0$\pm$0.1  \\  \hline
 \depth{5} & \depth{1} & 49.9$\pm$0.7 & 43.7$\pm$4.4 & 49.1$\pm$1.1  \\
 & \depth{2} & 49.1$\pm$0.3 & 43.4$\pm$3.9 & 51.4$\pm$0.6  \\
 & \depth{3} & 50.6$\pm$0.2 & 44.3$\pm$2.7 & 50.5$\pm$0.3  \\
 & \depth{4} & 50.9$\pm$0.8 & 44.4$\pm$3.4 & 50.3$\pm$0.4  \\
 & \depth{5} & 100.0$\pm$0.0 & 100.0$\pm$0.0 & 100.0$\pm$0.1  \\ \hline
 \end{tabular}
}
  \caption{Results for localism.}
 \label{tab:emb7_8}
\end{table}
Table~\ref{tab:emb7_8} shows the performance of all models on localism of embedding monotonicity.
When the models were trained with \depth{3}, \depth{4} or \depth{5},
all performed at around chance on the test set of non-embedding inferences \depth{1} and the test set of inferences involving one embedded clause \depth{2}.
These results indicate that even if models are trained with a set of inferences containing complex syntactic structures,
the models fail to locally interpret their constituents.

\begin{table}[tb]
\centering
\scalebox{0.7}{
\begin{tabular}[t]{llccc} \hline
\textbf{Train} & \textbf{Dev/Test} & \textbf{BERT} & \textbf{LSTM} & \textbf{TreeLSTM} \\ \hline
MNLI&\depth{1} & 46.9$\pm$0.4 & 47.2$\pm$1.1 & 43.4$\pm$0.3  \\
 & \depth{2} & 46.2$\pm$0.6 & 48.3$\pm$1.0 & 49.5$\pm$0.4  \\
 & \depth{3} & 46.8$\pm$0.8 & 48.9$\pm$0.7 & 41.0$\pm$0.4  \\
 & \depth{4} & 48.5$\pm$0.8 & 50.6$\pm$0.5 & 48.5$\pm$0.2  \\
 & \depth{5} & 48.9$\pm$0.6 & 49.3$\pm$0.7 & 48.8$\pm$0.5  \\
 &MNLI-test&84.6$\pm$0.2&64.7$\pm$0.3&70.4$\pm$0.1\\ \hline
$\depth{1}\adddata\depth{2}$& \depth{1} & 100.0$\pm$0.0 & 100.0$\pm$0.1 & 100.0$\pm$0.1  \\
 $\adddata$MNLI& \depth{2} & 100.0$\pm$0.0 & 89.3$\pm$9.0 & 99.8$\pm$0.1  \\
 & \depth{3} & 67.8$\pm$12.5 & 66.7$\pm$13.5 & 76.3$\pm$4.1  \\
 & \depth{4} & 46.8$\pm$3.7 & 47.1$\pm$14.6 & 50.7$\pm$7.8  \\
 & \depth{5} & 41.2$\pm$4.3 & 46.7$\pm$11.2 & 47.5$\pm$3.7  \\
 &MNLI-test & 84.4$\pm$0.2 & 39.7$\pm$0.5 & 63.0$\pm$0.2  
 \\\hline
$\depth{1}\adddata\depth{2}\adddata\depth{3}$ & \depth{1} & 100.0$\pm$0.0 & 100.0$\pm$0.0 & 100.0$\pm$0.0  \\
 $\adddata$MNLI& \depth{2} & 100.0$\pm$0.0 & 97.1$\pm$5.0 & 99.8$\pm$0.0  \\
 & \depth{3} & 100.0$\pm$0.0 & 89.2$\pm$5.1 & 98.3$\pm$1.1  \\
 & \depth{4} & 70.9$\pm$7.9 & 73.4$\pm$10.9 & 76.1$\pm$5.6  \\
 & \depth{5} & 42.4$\pm$4.2 & 47.8$\pm$3.9 & 57.0$\pm$4.3  \\
 &MNLI-test & 84.0$\pm$0.1 & 39.7$\pm$0.4 & 62.8$\pm$0.2  \\
 \hline
 \end{tabular}
 }
  \caption{Results for productivity where models were trained with our synthesized dataset mixed with MultiNLI (MNLI).}
 \label{tab:mnliemb1_2}
\end{table}

\paragraph{\todo{Performance of data augmentation}}
Prior studies~\cite{yanaka2019, Richardson2019} have shown that given BERT initially trained with MultiNLI, further training with synthesized instances of logical inference improves performance on the same types of logical inference while maintaining the initial performance on MultiNLI.
To investigate whether the results of our study are transferable to current work on MultiNLI,
we trained models with our synthesized dataset mixed with MultiNLI, and checked (i)~whether our synthesized dataset degrades the original performance of models on MultiNLI\footnote{Following the previous work~\cite{Richardson2019}, we used the MultiNLI mismatched development set for MNLI-test.} and (ii)~whether MultiNLI degrades the ability to generalize to unseen depths of embedded clauses.

Table~\ref{tab:mnliemb1_2} shows that training BERT on our synthetic data $\depth{1}\adddata\depth{2}$ and MultiNLI increases the accuracy on our test sets $\depth{1}$ (46.9 to 100.0), $\depth{2}$ (46.2 to 100.0), and $\depth{3}$ (46.8 to 67.8) while preserving accuracy on MultiNLI (84.6 to 84.4).
This indicates that training BERT with our synthetic data does not degrade performance on commonly used corpora like MultiNLI while improving the performance on monotonicity, which suggests that our data-synthesis approach can be combined with naturalistic datasets.
For TreeLSTM and LSTM, however, adding our synthetic dataset decreases accuracy on MultiNLI.
One possible reason for this is that a pre-training based model like BERT can mitigate catastrophic forgetting in various types of datasets.

Regarding the ability to generalize to unseen depths of embedded clauses, the accuracy of all models on our synthetic test set containing embedded clauses one level deeper than the training set exceeds chance, but the improvement becomes smaller with the addition of MultiNLI.
In particular, with the addition of MultiNLI, the models tend to change wrong predictions in cases where a hypothesis contains a phrase not occurring in a premise but the premise entails the hypothesis.
Such inference patterns are contrary to the heuristics in MultiNLI~\cite{mccoy2019}.
This indicates that there may be some trade-offs in terms of performance between inference patterns in the training set and those in the test set.

\section{Related Work}
The question of whether neural networks are capable of processing compositionality has been widely discussed~\cite{Fodor1988-FODCAC,962dc7dfb35547148019f194381d2cc6}.
Recent empirical studies illustrate the importance and difficulty of evaluating the capability of neural models.
Generation tasks using artificial datasets have been proposed for testing whether models compositionally interpret training data
from the underlying grammar of the data~\cite{Lake2017GeneralizationWS,hupkes2018,saxton2018analysing,loula-etal-2018-rearranging,hupkes2019,Bernardy2018CanRN}.
However, these conclusions are controversial, and it remains unclear whether the failure of models on these tasks stems from their inability to deal with compositionality.

Previous studies using logical inference tasks have also reported both positive and negative results.
Assessment results on propositional logic~\cite{Evans2018CanNN}, first-order logic~\cite{mul2019}, and natural logic~\cite{Bowman2015} show that neural networks can generalize to unseen words and lengths.
In contrast, \citet{Geiger2019} obtained negative results by testing models under fair conditions of natural logic.
Our study suggests that these conflicting results come from an absence of perspective on combinations of semantic properties.

Regarding assessment of the behavior of modern language models, \citet{linzen-etal-2016-assessing}, \citet{tran-etal-2018-importance}, and \citet{goldberg2019} investigated their syntactic capabilities by testing such models on subject--verb agreement tasks.
Many studies of NLI tasks~\cite{liu-etal-2019-inoculation,glockner-shwartz-goldberg:2018:Short,poliak-EtAl:2018:S18-2,DBLP:conf/lrec/Tsuchiya18,mccoy2019,rozen-etal-2019-diversify,ross-pavlick-2019-well} have provided evaluation methodologies and found that current NLI models often fail on particular inference types, or that they learn undesired heuristics from the training set.
\todo{In particular, recent works~\cite{yanaka-etal-2019-neural,yanaka2019,Richardson2019} have evaluated models on monotonicity, but did not focus on the ability to generalize to unseen combinations of patterns.}
Monotonicity covers various systematic inferential patterns, and thus is an adequate semantic phenomenon for assessing inferential systematicity in natural language.
Another benefit of focusing on monotonicity is that it provides hard problem settings against heuristics~\citep{mccoy2019}, which fail to perform downward-entailing inferences where the hypothesis is longer than the premise.

\section{Conclusion}
\label{sec:conc}
\todo{We introduced a method for evaluating whether DNN-based models can learn systematicity of monotonicity inference under four aspects.}
A series of experiments showed that the capability of three models to capture systematicity of predicate replacements was limited to cases where the positions of the constituents were similar between the training and test sets.
For embedding monotonicity,
no models consistently drew inferences involving embedded clauses whose depths were two levels deeper than those in the training set.
This suggests that models fail to capture inferential systematicity of monotonicity and its productivity.

\todo{We also found that BERT trained with our synthetic dataset mixed with MultiNLI maintained performance on MultiNLI while improving the performance on monotonicity.
This indicates that though current DNN-based models do not systematically interpret monotonicity inference, some models might have sufficient ability to memorize different types of reasoning.}
We hope that our work will be useful in future research for realizing more advanced models that are capable of appropriately performing arbitrary inferences.

\section*{Acknowledgement}
We thank the three anonymous reviewers for their helpful comments and suggestions.
We are also grateful to Benjamin Heinzerling and Sosuke Kobayashi for helpful discussions.
This work was partially supported by JSPS KAKENHI Grant Numbers JP20K19868 and JP18H03284, Japan.

\bibliographystyle{acl_natbib}
\bibliography{acl2020}

\begin{thebibliography}{41}
\expandafter\ifx\csname natexlab\endcsname\relax\def\natexlab#1{#1}\fi

\bibitem[{Abzianidze(2015)}]{abzianidze:2015:EMNLP}
Lasha Abzianidze. 2015.
\newblock A tableau prover for natural logic and language.
\newblock In \emph{Proceedings of the 2015 Conference on Empirical Methods in
  Natural Language Processing (EMNLP-2015)}, pages 2492--2502.

\bibitem[{Aydede(1997)}]{Aydede1997}
Murat Aydede. 1997.
\newblock Language of thought: The connectionist contribution.
\newblock \emph{Minds and Machines}, 7(1):57--101.

\bibitem[{van Benthem(1983)}]{10.2307/25001141}
Johan van Benthem. 1983.
\newblock Determiners and logic.
\newblock \emph{Linguistics and Philosophy}, 6(4):447--478.

\bibitem[{Bernardy(2018)}]{Bernardy2018CanRN}
Jean-Philippe Bernardy. 2018.
\newblock Can recurrent neural networks learn nested recursion.
\newblock \emph{Linguistic Issues in Language Technology}, 16.

\bibitem[{Blackburn and Bos(2005)}]{BlackburnBos05}
Patrick Blackburn and Johan Bos. 2005.
\newblock \emph{Representation and Inference for Natural Language: A First
  Course in Computational Semantics}.
\newblock Center for the Study of Language and Information.

\bibitem[{Bowman et~al.(2015)Bowman, Potts, and Manning}]{Bowman2015}
Samuel~R. Bowman, Christopher Potts, and Christopher~D. Manning. 2015.
\newblock Recursive neural networks can learn logical semantics.
\newblock In \emph{Proceedings of the 3rd Workshop on Continuous Vector Space
  Models and their Compositionality}, pages 12--21.

\bibitem[{Dagan et~al.(2013)Dagan, Roth, Sammons, and
  Zanzotto}]{series/synthesis/2013Dagan}
Ido Dagan, Dan Roth, Mark Sammons, and Fabio~Massimo Zanzotto. 2013.
\newblock \emph{Recognizing Textual Entailment: Models and Applications}.
\newblock Morgan \& Claypool Publishers.

\bibitem[{Devlin et~al.(2019)Devlin, Ming-Wei, Kenton, and
  Kristina}]{BERT2018new}
Jacob Devlin, Chang Ming-Wei, Lee Kenton, and Toutanova Kristina. 2019.
\newblock {BERT}: Pre-training of deep bidirectional transformers for language
  understanding.
\newblock In \emph{Proceedings of the 2019 Conference of the North American
  Chapter of the Association for Computational Linguistics: Human Language
  Technologies (NAACL-HLT-2019)}, pages 4171--4186.

\bibitem[{Evans et~al.(2018)Evans, Saxton, Amos, Kohli, and
  Grefenstette}]{Evans2018CanNN}
Richard Evans, David Saxton, David Amos, Pushmeet Kohli, and Edward
  Grefenstette. 2018.
\newblock Can neural networks understand logical entailment?
\newblock In \emph{International Conference on Learning Representations
  (ICLR-2018)}.

\bibitem[{Fodor and Pylyshyn(1988)}]{Fodor1988-FODCAC}
Jerry~A. Fodor and Zenon~W. Pylyshyn. 1988.
\newblock Connectionism and cognitive architecture: A critical analysis.
\newblock \emph{Cognition}, 28(1-2):3--71.

\bibitem[{Geiger et~al.(2019)Geiger, Cases, Karttunen, and Potts}]{Geiger2019}
Atticus Geiger, Ignacio Cases, Lauri Karttunen, and Christopher Potts. 2019.
\newblock Posing fair generalization tasks for natural language inference.
\newblock In \emph{Proceedings of the 2019 Conference on Empirical Methods in
  Natural Language Processing and the 9th International Joint Conference on
  Natural Language Processing (EMNLP-IJCNLP-2019)}, pages 4484--4494.

\bibitem[{Glockner et~al.(2018)Glockner, Shwartz, and
  Goldberg}]{glockner-shwartz-goldberg:2018:Short}
Max Glockner, Vered Shwartz, and Yoav Goldberg. 2018.
\newblock Breaking {NLI} systems with sentences that require simple lexical
  inferences.
\newblock In \emph{Proceedings of the 56th Annual Meeting of the Association
  for Computational Linguistics (ACL-2018)}, pages 650--655.

\bibitem[{Goldberg(2019)}]{goldberg2019}
Yoav Goldberg. 2019.
\newblock Assessing {BERT}'{}s syntactic abilities.
\newblock \emph{CoRR}, abs/1901.05287.

\bibitem[{Hochreiter and
  Schmidhuber(1997)}]{Hochreiter:1997:LSM:1246443.1246450}
Sepp Hochreiter and J\"{u}rgen Schmidhuber. 1997.
\newblock Long short-term memory.
\newblock \emph{Neural Comput.}, 9(8):1735--1780.

\bibitem[{Hu et~al.(2019)Hu, Chen, Richardson, Mukherjee, Moss, and
  K{\"u}bler}]{Hu2019MonaLogAL}
Hai Hu, Qi~Chen, Kyle Richardson, Atreyee Mukherjee, Lawrence~S. Moss, and
  Sandra K{\"u}bler. 2019.
\newblock Monalog: a lightweight system for natural language inference based on
  monotonicity.
\newblock \emph{CoRR}, abs/1910.08772.

\bibitem[{Hupkes et~al.(2019)Hupkes, Dankers, Mul, and Bruni}]{hupkes2019}
Dieuwke Hupkes, Verna Dankers, Mathijs Mul, and Elia Bruni. 2019.
\newblock Compositionality decomposed: how do neural networks generalise?
\newblock \emph{CoRR}, abs/1908.08351.

\bibitem[{Hupkes et~al.(2018)Hupkes, Veldhoen, and Zuidema}]{hupkes2018}
Dieuwke Hupkes, Sara Veldhoen, and Willem Zuidema. 2018.
\newblock Visualisation and `diagnostic classifiers' reveal how recurrent and
  recursive neural networks process hierarchical structure.
\newblock \emph{Journal of Artificial Intelligence Research}, 61:907--926.

\bibitem[{Icard and Moss(2014)}]{moss2014}
Thomas Icard and Lawrence Moss. 2014.
\newblock Recent progress in monotonicity.
\newblock \emph{Linguistic Issues in Language Technology}, 9(7):167--194.

\bibitem[{Kingma and Ba(2015)}]{Adam}
Diederik~P. Kingma and Jimmy Ba. 2015.
\newblock Adam: A method for stochastic optimization.
\newblock In \emph{International Conference on Learning Representations
  (ICLR-2015)}.

\bibitem[{Lake and Baroni(2017)}]{Lake2017GeneralizationWS}
Brenden~M. Lake and Marco Baroni. 2017.
\newblock Generalization without systematicity: On the compositional skills of
  sequence-to-sequence recurrent networks.
\newblock In \emph{International Conference on Machine Learning (ICML-2017)}.

\bibitem[{Linzen et~al.(2016)Linzen, Dupoux, and
  Goldberg}]{linzen-etal-2016-assessing}
Tal Linzen, Emmanuel Dupoux, and Yoav Goldberg. 2016.
\newblock Assessing the ability of {LSTM}s to learn syntax-sensitive
  dependencies.
\newblock \emph{Transactions of the Association for Computational Linguistics},
  4:521--535.

\bibitem[{Liu et~al.(2019)Liu, Schwartz, and Smith}]{liu-etal-2019-inoculation}
Nelson~F. Liu, Roy Schwartz, and Noah~A. Smith. 2019.
\newblock Inoculation by fine-tuning: A method for analyzing challenge
  datasets.
\newblock In \emph{Proceedings of the 2019 Conference of the North {A}merican
  Chapter of the Association for Computational Linguistics: Human Language
  Technologies (NAACL-HLT-2019)}, pages 2171--2179.

\bibitem[{Loula et~al.(2018)Loula, Baroni, and
  Lake}]{loula-etal-2018-rearranging}
Jo{\~a}o Loula, Marco Baroni, and Brenden Lake. 2018.
\newblock Rearranging the familiar: Testing compositional generalization in
  recurrent networks.
\newblock In \emph{Proceedings of the 2018 {EMNLP} Workshop {B}lackbox{NLP}:
  Analyzing and Interpreting Neural Networks for {NLP}}, pages 108--114.

\bibitem[{Marcus(2003)}]{962dc7dfb35547148019f194381d2cc6}
Gary Marcus. 2003.
\newblock \emph{The Algebraic Mind: Integrating Connectionism and Cognitive
  Science}.
\newblock MIT Press.

\bibitem[{McCoy et~al.(2019)McCoy, Pavlick, and Linzen}]{mccoy2019}
R.~Thomas McCoy, Ellie Pavlick, and Tal Linzen. 2019.
\newblock Right for the wrong reasons: Diagnosing syntactic heuristics in
  natural language inference.
\newblock In \emph{Proceedings of the 57th Annual Meeting of the Association
  for Computational Linguistics (ACL-2019)}, pages 3428--3448.

\bibitem[{Mineshima et~al.(2015)Mineshima, Mart\'{i}nez-G\'{o}mez, Miyao, and
  Bekki}]{D16-1242}
Koji Mineshima, Pascual Mart\'{i}nez-G\'{o}mez, Yusuke Miyao, and Daisuke
  Bekki. 2015.
\newblock Higher-order logical inference with compositional semantics.
\newblock In \emph{Proceedings of the 2015 Conference on Empirical Methods in
  Natural Language Processing (EMNLP-2015)}, pages 2055--2061.

\bibitem[{Mul and Zuidema(2019)}]{mul2019}
Mathijs Mul and Willem Zuidema. 2019.
\newblock Siamese recurrent networks learn first-order logic reasoning and
  exhibit zero-shot compositional generalization.
\newblock \emph{CoRR}, abs/1908.08351.

\bibitem[{Pagin and Westerst\aa{}hl(2010)}]{Pagin2010-PAGCID}
Peter Pagin and Dag Westerst\aa{}hl. 2010.
\newblock Compositionality {I}: Definitions and variants.
\newblock \emph{Philosophy Compass}, 5(3):250--264.

\bibitem[{Pennington et~al.(2014)Pennington, Socher, and
  Manning}]{pennington-etal-2014-glove}
Jeffrey Pennington, Richard Socher, and Christopher Manning. 2014.
\newblock {G}love: Global vectors for word representation.
\newblock In \emph{Proceedings of the 2014 Conference on Empirical Methods in
  Natural Language Processing ({EMNLP}-2014)}, pages 1532--1543.

\bibitem[{Poliak et~al.(2018)Poliak, Naradowsky, Haldar, Rudinger, and
  Van~Durme}]{poliak-EtAl:2018:S18-2}
Adam Poliak, Jason Naradowsky, Aparajita Haldar, Rachel Rudinger, and Benjamin
  Van~Durme. 2018.
\newblock Hypothesis only baselines in natural language inference.
\newblock In \emph{Proceedings of the Seventh Joint Conference on Lexical and
  Computational Semantics (*SEM-2018)}, pages 180--191.

\bibitem[{Richardson et~al.(2020)Richardson, Hu, Moss, and
  Sabharwal}]{Richardson2019}
Kyle Richardson, Hai Hu, Lawrence~S. Moss, and Ashish Sabharwal. 2020.
\newblock Probing natural language inference models through semantic fragments.
\newblock In \emph{Proceedings of the 34th AAAI Conference on Artificial
  Intelligence (AAAI-2020)}.

\bibitem[{Ross and Pavlick(2019)}]{ross-pavlick-2019-well}
Alexis Ross and Ellie Pavlick. 2019.
\newblock How well do {NLI} models capture verb veridicality?
\newblock In \emph{Proceedings of the 2019 Conference on Empirical Methods in
  Natural Language Processing and the 9th International Joint Conference on
  Natural Language Processing (EMNLP-IJCNLP-2019)}, pages 2230--2240.

\bibitem[{Rozen et~al.(2019)Rozen, Shwartz, Aharoni, and
  Dagan}]{rozen-etal-2019-diversify}
Ohad Rozen, Vered Shwartz, Roee Aharoni, and Ido Dagan. 2019.
\newblock Diversify your datasets: Analyzing generalization via controlled
  variance in adversarial datasets.
\newblock In \emph{Proceedings of the 23rd Conference on Computational Natural
  Language Learning (CoNLL-2019)}, pages 196--205.

\bibitem[{Saxton et~al.(2019)Saxton, Grefenstette, Hill, and
  Kohli}]{saxton2018analysing}
David Saxton, Edward Grefenstette, Felix Hill, and Pushmeet Kohli. 2019.
\newblock Analysing mathematical reasoning abilities of neural models.
\newblock In \emph{International Conference on Learning Representations
  (ICLR-2019)}.

\bibitem[{Tran et~al.(2018)Tran, Bisazza, and Monz}]{tran-etal-2018-importance}
Ke~Tran, Arianna Bisazza, and Christof Monz. 2018.
\newblock The importance of being recurrent for modeling hierarchical
  structure.
\newblock In \emph{Proceedings of the 2018 Conference on Empirical Methods in
  Natural Language Processing (EMNLP-2018)}, pages 4731--4736.

\bibitem[{Tran and Cheng(2018)}]{tran-cheng-2018-multiplicative}
Nam~Khanh Tran and Weiwei Cheng. 2018.
\newblock Multiplicative tree-structured long short-term memory networks for
  semantic representations.
\newblock In \emph{Proceedings of the Seventh Joint Conference on Lexical and
  Computational Semantics (*SEM-2018)}, pages 276--286.

\bibitem[{Tsuchiya(2018)}]{DBLP:conf/lrec/Tsuchiya18}
Masatoshi Tsuchiya. 2018.
\newblock Performance impact caused by hidden bias of training data for
  recognizing textual entailment.
\newblock In \emph{Proceedings of the 11th International Conference on Language
  Resources and Evaluation (LREC-2018)}.

\bibitem[{Wang et~al.(2019)Wang, Singh, Michael, Hill, Levy, and
  Bowman}]{wang2018glue}
Alex Wang, Amanpreet Singh, Julian Michael, Felix Hill, Omer Levy, and Samuel
  Bowman. 2019.
\newblock {GLUE}: A multi-task benchmark and analysis platform for natural
  language understanding.
\newblock In \emph{Proceedings of the International Conference on Learning
  Representations (ICLR-2019)}.

\bibitem[{Williams et~al.(2018)Williams, Nangia, and
  Bowman}]{DBLP:journals/corr/WilliamsNB17}
Adina Williams, Nikita Nangia, and Samuel Bowman. 2018.
\newblock A broad-coverage challenge corpus for sentence understanding through
  inference.
\newblock In \emph{Proceedings of the 2018 Conference of the North American
  Chapter of the Association for Computational Linguistics: Human Language
  Technologies (NAACL-HLT-2018)}, pages 1112--1122.

\bibitem[{Yanaka et~al.(2019{\natexlab{a}})Yanaka, Mineshima, Bekki, Inui,
  Sekine, Abzianidze, and Bos}]{yanaka-etal-2019-neural}
Hitomi Yanaka, Koji Mineshima, Daisuke Bekki, Kentaro Inui, Satoshi Sekine,
  Lasha Abzianidze, and Johan Bos. 2019{\natexlab{a}}.
\newblock Can neural networks understand monotonicity reasoning?
\newblock In \emph{Proceedings of the 2019 ACL Workshop BlackboxNLP: Analyzing
  and Interpreting Neural Networks for NLP}, pages 31--40.

\bibitem[{Yanaka et~al.(2019{\natexlab{b}})Yanaka, Mineshima, Bekki, Inui,
  Sekine, Abzianidze, and Bos}]{yanaka2019}
Hitomi Yanaka, Koji Mineshima, Daisuke Bekki, Kentaro Inui, Satoshi Sekine,
  Lasha Abzianidze, and Johan Bos. 2019{\natexlab{b}}.
\newblock {HELP}: A dataset for identifying shortcomings of neural models in
  monotonicity reasoning.
\newblock In \emph{Proceedings of the Eighth Joint Conference on Lexical and
  Computational Semantics (*{SEM} 2019)}, pages 250--255.

\end{thebibliography}

\appendix

\begin{table*}[bt!]
    \begin{tabular}{lll}\\ \hline
    \multicolumn{3}{c}{\textbf{Context-free grammar for premise sentences}} \\
    $S$ & $\rightarrow$& $NP \ \, IV_1$\\
    $\mathit{NP}$ & $\rightarrow$ & $Q \ \, N$ \ $\mid$ \ $Q \ \, N \ \, \overline{S}$\\
    $\overline{S}$ & $\rightarrow$ & $\mathit{WhNP}\ \, TV\ \, NP \mid \mathit{WhNP}\ \, \mathit{NP}\ \, TV \mid \mathit{NP}\ \, TV$ \\ \hline
    \multicolumn{3}{c}{\textbf{Lexicon}} \\ 
    $Q$&$\rightarrow$&\{\textit{no, at most three, less than three, few, some, at least three, more than three, a few}\}\\
    $N$&$\rightarrow$&\{\textit{dog, rabbit, lion, cat, bear, tiger, elephant, fox, monkey, wolf}\}\\
    $IV_1$&$\rightarrow$&\{\textit{ran, walked, came, waltzed, swam, rushed, danced, dawdled, escaped, left}\}\\
    $IV_2$&$\rightarrow$&\{\textit{laughed, groaned, roared, screamed, cried}\}\\
    $TV$&$\rightarrow$&\{\textit{kissed, kicked, hit, cleaned, touched, loved, accepted, hurt, licked, followed}\}\\
    $WhNP$&$\rightarrow$&\{\textit{that, which}\}\\ 
    $N_{hypn}$&$\rightarrow$&\{\textit{animal, creature, mammal, beast}\}\\
    $Adj$&$\rightarrow$&\{\textit{small, large, crazy, polite, wild}\}\\
    $PP$&$\rightarrow$&\{\textit{in the area, on the ground, at the park, near the shore, around the island}\}\\
    $RelC$&$\rightarrow$&\{\textit{which ate dinner, that liked flowers, which hated the sun, that stayed up late}\}\\
    $Adv$&$\rightarrow$&\{\textit{slowly, quickly, seriously, suddenly, lazily}\}\\ \hline
    \multicolumn{3}{c}{\textbf{Predicate replacements for hypothesis sentences}} \\
    $N$ & to & $N_{hypn} \mid Adj\ N \mid N\ PP \mid N\ RelC$\\
    $IV_1$ & to & $IV_1\ Adv \mid IV_1\ PP \mid IV_1 \ \text{or} \ IV_2 \mid IV_1 \ \text{and} \ IV_2$\\ \hline

    \end{tabular}
    \caption{A context-free grammar and a set of predicate replacements used to generate inference examples. Predicate replacement is applied to $N$ or $IV_1$, replacing it with a corresponding phrase.}
    \label{tab:lexicon}
\end{table*}

\section{Appendix}
\subsection{Lexical entries and replacement examples}
Table~\ref{tab:lexicon} shows a context-free grammar and a set of predicate replacements used to generate inference examples.
Regarding the context-free grammar, we consider premise--hypothesis pairs containing the quantifier $Q$ in the subject position, and the predicate replacement is performed in both the first and second arguments of the quantifier.
When generating premise--hypothesis pairs involving embedding monotonicity, we 
consider inferences involving four types of predicate replacements (hyponyms $N_{hypn}$, adjectives $Adj$, prepositions $PP$, and relative clauses $RelC$) in the first argument of the quantifier.
To generate natural sentences consistently, we use the past tense for verbs;
for lexical entries and predicate replacements, we select those that do not violate selectional restriction.

To check the gold labels for the generated premise--hypothesis pairs, 
we translate each sentence to a first-order logic (FOL) formula
and test if the entailment relation holds by theorem proving.
The FOL formulas are
compositionally derived by combining lambda terms assigned to each lexical item in accordance with meaning composition rules specified in the CFG rules in the standard way~\cite{BlackburnBos05}.
Since our purpose is to check the polarity of monotonicity marking, 
vague quantifiers such as \textit{few} are represented according to their polarity.
For example, we map the quantifier \textit{few} onto the lambda-term $\lambda P \lambda Q \lnot \exists x (\textbf{few}(x) \land P(x) \wedge Q(x))$.

\subsection{Results on embedding monotonicity}
Table~\ref{tab:emball} shows all results on embedding monotonicity.
\todo{This indicates that all models partially generalize to inferences containing embedded clauses one level deeper than the training set, but fail to generalize to inferences containing embedded clauses two or more levels deeper.}

\begin{table*}[tb]
\centering
\begin{tabular}[t]{llccc} \hline
\textbf{Train} & \textbf{Test} & \textbf{BERT} & \textbf{LSTM} & \textbf{TreeLSTM} \\ \hline
\depth{1} & \depth{1} & 100.0$\pm$0.0 & 91.1$\pm$5.4 & 100.0$\pm$0.0 \\
 & \depth{2} & 44.1$\pm$6.4 & 34.1$\pm$3.8 & 48.1$\pm$1.2 \\
 & \depth{3} & 47.6$\pm$3.2 & 45.1$\pm$5.1 & 48.5$\pm$1.8  \\
 & \depth{4} & 49.6$\pm$1.0 & 44.4$\pm$6.5 & 50.1$\pm$2.1  \\
 & \depth{5} & 49.9$\pm$1.1 & 44.1$\pm$5.3 & 50.3$\pm$1.1  \\ \hline
$\depth{1}\cup\depth{2}$ & \depth{1} & 100.0$\pm$0.0 & 100.0$\pm$0.0 & 100.0$\pm$0.1  \\
 & \depth{2} & 100.0$\pm$0.0 & 99.8$\pm$0.2 & 99.5$\pm$0.1  \\
 & \depth{3} & 75.2$\pm$10.0 & 75.4$\pm$10.8 & 86.4$\pm$4.1  \\
 & \depth{4} & 55.0$\pm$3.7 & 57.7$\pm$8.7 & 58.6$\pm$7.8  \\
 & \depth{5} & 49.9$\pm$4.4 & 45.8$\pm$4.0 & 48.4$\pm$3.7  \\ \hline
$\depth{1}\cup\depth{2}\cup\depth{3}$ & \depth{1} & 100.0$\pm$0.0 & 100.0$\pm$0.0 & 100.0$\pm$0.0  \\
 & \depth{2} & 100.0$\pm$0.0 & 95.1$\pm$7.8 & 99.6$\pm$0.0  \\
 & \depth{3} & 100.0$\pm$0.0 & 85.2$\pm$8.9 & 97.7$\pm$1.1  \\
 & \depth{4} & 77.9$\pm$10.8 & 59.7$\pm$10.8 & 68$\pm$5.6  \\
 & \depth{5} & 53.5$\pm$19.6 & 55.1$\pm$8.2 & 49.6$\pm$4.3  \\ \hline
$\depth{1}\cup\depth{2}\cup\depth{3}\cup\depth{4}$ & \depth{1} & 100.0$\pm$0.0 & 100.0$\pm$0.0 & 100.0$\pm$0.1  \\
 & \depth{2} & 100.0$\pm$0.0 & 99.4$\pm$1.1 & 99.7$\pm$0.2  \\
 & \depth{3} & 100.0$\pm$0.0 & 91.5$\pm$4.0 & 98.9$\pm$1.1  \\
 & \depth{4} & 100.0$\pm$0.0 & 74.1$\pm$4.2 & 94.0$\pm$2.3  \\
 & \depth{5} & 89.1$\pm$5.4 & 64.2$\pm$4.7 & 69.5$\pm$4.1  \\ \hline
$\depth{1}\cup\depth{2}\cup\depth{3}\cup\depth{4}\cup\depth{5}$ & \depth{1} & 100.0$\pm$0.0 & 100.0$\pm$0.0 & 100.0$\pm$0.1  \\
 & \depth{2} & 100.0$\pm$0.0 & 95.8$\pm$7.3 & 99.8$\pm$0.1  \\
 & \depth{3} & 100.0$\pm$0.0 & 90.5$\pm$13.1 & 99.1$\pm$0.2  \\
 & \depth{4} & 100.0$\pm$0.0 & 90.2$\pm$6.0 & 94.8$\pm$0.1  \\
 & \depth{5} & 100.0$\pm$0.0 & 93.6$\pm$3.1 & 83.2$\pm$12.1  \\ \hline
\depth{2} & \depth{1} & 36.4$\pm$14.4 & 25.3$\pm$9.3 & 44.9$\pm$4.1  \\
 & \depth{2} & 100.0$\pm$0.0 & 100.0$\pm$0.0 & 100.0$\pm$0.2  \\
 & \depth{3} & 47.6$\pm$10.3 & 43.9$\pm$17.5 & 51.8$\pm$1.1  \\
 & \depth{4} & 61.7$\pm$7.8 & 57.9$\pm$14.7 & 51.7$\pm$0.6  \\
 & \depth{5} & 42.6$\pm$5.1 & 47.2$\pm$2.9 & 50.9$\pm$0.4  \\ \hline
\depth{3} & \depth{1} & 49.6$\pm$0.5 & 48.8$\pm$13.2 & 49.8$\pm$4.1  \\
 & \depth{2} & 49.8$\pm$0.6 & 47.3$\pm$12.1 & 51.8$\pm$1.1  \\
 & \depth{3} & 100.0$\pm$0.0 & 100.0$\pm$0.0 & 100.0$\pm$0.2  \\
 & \depth{4} & 49.7$\pm$1.0 & 42.0$\pm$0.6 & 51.3$\pm$0.7  \\
 & \depth{5} & 50.0$\pm$0.4 & 38.4$\pm$9.6 & 49.8$\pm$0.3  \\ \hline
\depth{4} & \depth{1} & 50.3$\pm$1.0 & 46.8$\pm$6.5 & 49.0$\pm$0.4  \\
 & \depth{2} & 49.6$\pm$0.8 & 45.4$\pm$1.8 & 49.7$\pm$0.3  \\
 & \depth{3} & 50.2$\pm$0.7 & 45.1$\pm$0.6 & 50.5$\pm$0.7  \\
 & \depth{4} & 100.0$\pm$0.0 & 100.0$\pm$0.0 & 100.0$\pm$0.1  \\
 & \depth{5} & 49.7$\pm$0.5 & 45.1$\pm$0.9 & 50.5$\pm$1.1  \\ \hline
\depth{5} & \depth{1} & 49.9$\pm$0.7 & 43.7$\pm$4.4 & 49.1$\pm$1.1  \\
 & \depth{2} & 49.1$\pm$0.3 & 43.4$\pm$3.9 & 51.4$\pm$0.6  \\
 & \depth{3} & 50.6$\pm$0.2 & 44.3$\pm$2.7 & 50.5$\pm$0.3  \\
 & \depth{4} & 50.9$\pm$0.8 & 44.4$\pm$3.4 & 50.3$\pm$0.4  \\
 & \depth{5} & 100.0$\pm$0.0 & 100.0$\pm$0.0 & 100.0$\pm$0.1  \\ \hline
 \end{tabular}
  \caption{All results on embedding monotonicity.}
 \label{tab:emball}
\end{table*}

\end{document}